\documentclass[letterpaper, 10 pt, conference]{ieeeconf}  

\IEEEoverridecommandlockouts                              
\usepackage{multirow}
\usepackage{graphicx}
\graphicspath{ {./pics/} }
\usepackage{graphics} 
\usepackage{array}

\title{\LARGE \bf
Machine Learning with Clos Networks
}

\author{Timothy Whiting$^{1}$* and Thiam Khean Hah$^{2}$* \\ Intel
\thanks{*Intel}
\thanks{$^{1}$Tim Whiting is with Intel for an Internship and is from Brigham Young University
        {\tt\small tim at studentbody.byu.edu}}%
\thanks{$^{2}$Thiam Khean Hah is with Intel
        {\tt\small thiam.khean.hah at intel.com}}%
}

\begin{document}

\maketitle
\thispagestyle{empty}
\pagestyle{empty}

\begin{abstract}

We present a new methodology for improving the accuracy of small neural networks by applying the concept of a clos network to achieve maximum expression in a smaller network. We explore the design space to show that more layers is beneficial, given the same number of parameters. We also present findings on how the relu nonlinearity affects accuracy in separable networks. We present results on early work with Cifar-10 dataset.

\end{abstract}

\section{INTRODUCTION}

Deep learning has been show to do better than humans in classification of images. This requires a large amount of data to train on, but also novel deep neural network structures. The data center era has brought the data, and new structures have been created by Google and others to perform amazingly well on classification. The problem with the state of the art deep networks is that they require a lot of parameters and multiply-accumulates to get world class accuracy. This increases the time to train, as well as the latency of the network. 

Neural networks on FGPAs have been shown to reduce latency for classification by implementing the functions in custom hardware logic. This is important in areas such as automated driving in order to reduce response time of the system. In order to reduce the size impact on an FPGA to get lower latency and lower power new neural network configurations need to be researched and developed.
Our research focuses on how we can accomplish that by using smaller layers, while still achieving respectable accuracy on the Cifar-10 dataset.

\subsection{Convolutional Neural Networks (CNNs)}

CNNs are a development that have advanced the state of the art in image recognition tasks. The input to each layer of the network is a collection of feature maps. The output is also a collection of feature maps. A layer can be said to transform M features into N different features. The input to the entire network is a special case where the features correspond to the 3 color channels RGB. The transformation happens by a collection of N 3x3 kernels. Each kernel convolves a 3x3 pixel area across all input channels and produces 1 output pixel. Each of the N kernels are trained and learn to recognize different features in the input space. Periodically the stride of the kernels is increased to 2 skipping a pixel in each xy dimension. This allows later kernels to look at higher order features that cross the 3x3 boundary. While the size of the image map decreases, the number of kernels or output feature channels increases, to be able to represent additional higher order features. At the end a fully connected deep neural network layer is applied to determine based on what features are present in the image what the output class should be.

\subsection{Related Works}
Google has made large steps in making CNNs more efficient with their MobileNet v1 and MobileNet v2, these drastically reduced the computation and connections, while still achieving high accuracy. MobileNet decouples the xy convolution from the cross-channel convolution. They call this layer a depthwise separable convolution. The xy convolution is called depthwise convolution, and the convolution over the input channels is called a pointwise convolution since it operates on a single pixel or point at a time. This reduces the number of multiply-accumulates and parameters, enabling classification on mobile devices with low latency. MobileNet v2 uses a recent discovery called residual layers which groups layers in groups of two and adds the input of the group to the output of the group. This is called a shortcut connection and reduces the time to train, because the neural net only has to learn the difference between what is on the input and what is on the output, rather than the entire transformation function. 
ShuffleNet built upon these two by realizing that the pointwise convolution was now the portion that had the most multiply-accumulates and parameters. In order to reduce this, ShuffleNet groups the input channels into groups does a smaller convolution across each group in the pointwise layer (pointwise group convolutions), shuffles the output channels, and then does another pointwise group convolution on the new groups. ShuffleNet has fewer multiply-accumulates than MobileNet v1, and uses residual layers to outperform MobileNet with fewer parameters.
Interleaved Convolutions is another paper which explored the idea of mixing layers in depthwise groups. They did not separate out the depthwise from pointwise, instead only doing depthwise layers with a shuffle. 

\section{Xxnet Network}

\subsection{Clos Network}
ShuffleNet and Interleaved Convolutions did not explore the design space fully, they left some important concepts untried. The idea of grouping inputs and shuffling is analogous to a clos network. In a clos network every input can be routed to every output, given some bound on how many can do so simultaneously. It is the most efficient way to implement a crossbar for routing purposes in chip design. Applying the concept to pointwise grouping. In a neural network in order for every output to see the influence of every input with the minimum group size, there needs to be three shuffles and group convolution stages. Two group convolutions do not capture the full information. By creating a net with three shuffles and pointwise group convolutions, we show that we can increase accuracy with the same and fewer number of parameters than nets with more groups and less pointwise group convolutions.
\subsection{Relu Nonlinearity}
Mobilenet v1 applies a relu nonlinearity after each mini-layer. This decreases the expression of all of the inputs at the output of each full layer, by reducing the information that can be passed between the depthwise and pointwise stages. A full convolutional layer has outputs that depend on every input and only then does the nonlinearity get applied. In order to maximize transfer of knowledge between the depthwise and pointwise mini-layers we experiment with removing the relu unit from the depthwise mini-layer as well as from all mini-layers except the last mini-layer in each of our full layers.
\subsection{Xxnet basic Network Structure} 
The Xxnet network is based off of MobileNet v1 from the Tensorflow experiments repository. We chose MobileNet v1 instead of MobileNet v2 to prove the clos network concept before experimenting with residual layers. Each network has a single depthwise mini-layer, and multiple pointwise mini-layers with group convolutions. We ran experiments with shuffles in different locations between these mini-layers. We call the collection of depthwise and pointwise mini-layers a xxnet-layer. There are 13 xxnet-layers with the feature size and image sizes from the original Mobilenet v1. The default image preprocessor from the Mobilenet v1 repository was used to scale and crop the images.
To compare the clos network against similarly defined networks we created two networks with the same or more parameters, with larger group sizes, but fewer pointwise mini-layers.

\section{Experiments}

\subsection{Cifar-10 Dataset}

Cifar-10 is a collection of 60,000 images of which 50,000 are training and 10,000 are test. It contains 10 output classes. We ran our experiments with a batch size of 32 and an exponential learning rate decay with 2 epochs per decay and decay factor of 0.94. We ran all runs for 120 epochs, at which point our Mobilenet v1 runs had mostly saturated. Future work will explore the saturation point of the new networks
as well as whether there is some overfitting with using a large network like Mobilenet on the Cifar-10 dataset.

\subsection{Control Tests}
In order to compare our results to Mobilenet v1 we ran Mobilenet v1 on Cifar-10.
We also ran Mobilenet v1 without a relu on the depthwise mini-layer. 
It achieved better accuracy which informed our observations that a relu is detrimental between mini-layers.

\begin{table}[h]
\caption{Mobilenet v1 accuracy on Cifar-10}
\label{mobilenet_v1}
\begin{center}
\begin{tabular}{|c|c|}
\hline
All Relu & Only Pointwise Relu\\
\hline
90.59 & 91.04\\
\hline
\end{tabular}
\end{center}
\end{table}   

\subsection{Xxnet Tests}

There are three main versions of Xxnet that we created each with a few variations. 
Xxnet v1x\_3 o1 or the full clos network contains 3 pointwise mini-layers, each with the optimal group size to reduce the number of parameters. Optimal group size was calculated as taking the square root of the number of feature maps. In the case where there is no clean square root, for example, 128 feature maps, we split it into the nearest powers of two, so group sizes of 8 and 16. There are many combinations of the pointwise mini-layers that can be created with these group sizes. We used a pattern of small big small. In the 128 case that would be 8 16 8. However for the layers that grow; for example from 64 to 128 feature maps, or 128 to 256 there needed to be additional patterns. In the 64 to 128 case we chose a pattern of 16 16 8, and in the 128 to 256 case we can take the square root so 16 16 16. There is more room for experimentation to find which combination gives the network the best accuracy for fewest parameters. 

Growth happens by tiling the input feature maps before the first pointwise layer.

Xxnet v1x\_2 o2 contains 2 pointwise mini-layers with the first pointwise mini-layer having twice the number of features per group and the second layer having the required group size to match the number of parameters in a same layer of the v1x\_3 o1 network.

Xxnet v1x\_1 o3 contains a single pointwise mini-layer with close to 3 times the group size as the xxnet v1x\_3 o1 network. This ensures that we can compare the effect of the number of mini-layers with the same number of parameters. Because the number of features in the original Mobilenet v1 layer definitions are not divisible by 3, we started with a layer of 33 features, and then grew by factors of 2 in the same places as the original Mobilenet v1. We then chose the number of groups so that the parameter count was similar to the other two xxnet networks 

The variations on these networks were created by shuffling the channels between various layers, so the same group wouldn’t be seen twice in a row, ensuring maximum expression of the network. The candidate locations for shuffles were after each pointwise mini-layer, and after growing, since the depthwise layer is group agnostic. We shuffled an odd number of times, to switch up the groups from layer to layer, so the same group aren’t seen layer after layer. For the initial experiments we decided to not shuffle after the growth.
In addition to shuffling we also experimented with where to place the relu. We tried three variations. First a relu in every mini-layer. Second a relu in all pointwise mini-layers but not in the depthwise mini-layer. Third a relu only at the last pointwise mini-layer. Note that the second and third options are identical for a network with only one pointwise mini-layer.

\subsection{Equations}
The equation for the number of parameters in a mobilenet layer with M input channels, $D_k$ depthwise kernel size, and N output channels is:
$$
M*D_k^2 + M*N \label{mobilenet_eqn}
$$
The equation for the number of parameters in a given xxnet v1\_3 o1 layer with group sizes $G_1$ $G_2$ and $G_3$ is:
$$
M*D_k^2 + N/G_1*G_1^2+ N/G_2*G_2^2+N/G_3*G_3^2 \label{xxnet_v1_3_eqn}
$$
which reduces to:
$$
M*D_k^2 + N(G_1+G_2+G_3) \label{xxnet_v1_3_small_eqn}
$$
for an optimal group size of $G_1 = G_2 = G_3 = \sqrt[]{N}$ this is:
$$
M*D_k^2 + N(3*\sqrt[]{N}) \label{xxnet_v1_3_sqr_root_eqn}
$$

For a N of 64 and an M of 32 that is:

mobilenet: $32*9 + 32*64$

xxnet:     $32*9 + 24*64$ \\

For a larger layer such as from 512 to 1024 feature maps, this is:

mobilenet: $512*9 + 512*1024$

xxnet: $512*9 + 96*1024$\\

In total the parameter savings of xxnet over Mobilenet v1 is $>6.6x$ in the convolutional layers. 

\subsection{Parameter Counts}
\begin{table}[ht]
\caption{Parameter Count}
\label{param_count}
\begin{center}
\begin{tabular}{|c|c|c|c|c|}
\hline
Layer & Mobilenet v1 & v1x\_ 3 o1 & v1x\_ 2 o2 & v1x\_ 1 o3  \\
\hline
1 & 2336 & 1824 & 1824 & 2475 \\
\hline
2 & 8768 & 5696 & 5696 & 3498  \\
\hline
3 & 17536 & 5248 & 5248 & 6996 \\
\hline
4 & 33920 & 13440 & 13440 & 12804 \\
\hline
5 & 67840 & 14592 & 14592 & 13992 \\
\hline
6 & 133376 & 43264 & 43264 & 48840  \\
\hline
7 & 266752 & 37376 & 37376 & 51216  \\
\hline
8 & 266752 & 37376 & 37376 & 51216 \\
\hline
9 & 266752 & 37376 & 37376 & 51216  \\
\hline
10 & 266752 & 37376 & 37376 & 51216 \\
\hline
11 & 266752 & 37376 & 37376 & 51216 \\
\hline
12 & 528896 & 102912 & 102912 & 97680 \\
\hline
13 & 1057792 & 107520 & 107520 & 102432 \\
\hline
Total & 3184224 & 481376 & 481376 & 544797 \\
\hline
\end{tabular}
\end{center}
\end{table}
\section{Results}

Shuffles are shown across the top of each table. 000 indicates there is no shuffles. 00x indicates a shuffle after the last pointwise layer. 0x0 indicates a shuffle before the last pointwise layer, etc. 

\begin{table}[ht]
\caption{xxnet accuracy on cifar-10 with all relu}
\label{xxnet_all_relu}
\begin{center}
\begin{tabular}{|c|c|c|c|c|c|c|}
\hline
& Shuffles & 000 & 00x & 0x0 & 0xx & xxx\\
\hline
\multirow{3}{*}{\shortstack{Major\\ Network}} & v1x\_3 o1 & 74.83 & 81.13 & 83.57 & N/A & 82.97 \\
\cline{2-7}
& v1x\_2 o2 & 82.73 & 85.98 & 87.63 & 87.41 & N/A \\
\cline{2-7}
& v1x\_1 o3 & 87.81 & 88.85 & N/A & N/A & N/A \\
\hline
\end{tabular}
\end{center}
\end{table}
\begin{figure}[thpb]
      \centering
      \framebox{\parbox{3.2in}{
      \includegraphics[scale=0.5]{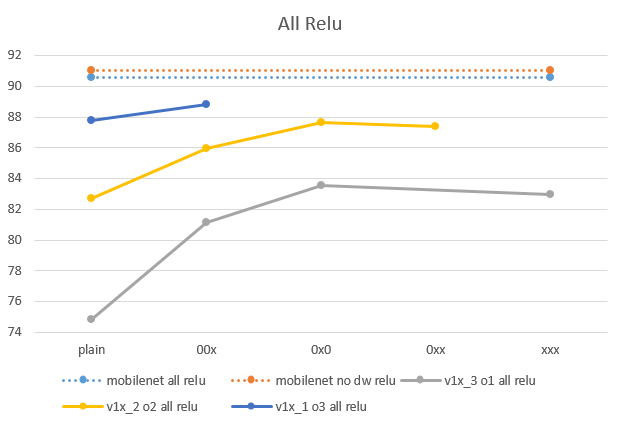}
      }}
      \label{allrelufig}
      \caption{}
\end{figure}

As can be seen in Table \ref{xxnet_all_relu} and Figure \ref{allrelufig} the networks did much worse than the baseline mobilenet, with fewer layers being better. Also the number of shuffles increase the accuracy, but only to a point.

\begin{table}[ht]
\caption{xxnet accuracy on cifar-10 with only pointwise relu}
\label{xxnet_no_dw_relu}
\begin{center}
\begin{tabular}{|c|c|c|c|c|c|c|}
\hline
& Shuffles & 000 & 00x & 0x0 & 0xx & xxx\\
\hline
\multirow{3}{*}{\shortstack{Major\\ Network}} & v1x\_3 o1 & 82.01 & 86.3 & 86.74 & N/A & 87.62 \\
\cline{2-7}
& v1x\_2 o2 & 85.89 & 88.99 & 89.79 & 89.5 & N/A \\
\cline{2-7}
& v1x\_1 o3 & 89.6 & 90.53 & N/A & N/A & N/A \\
\hline
\end{tabular}
\end{center}
\end{table}
\begin{figure}[thpb]
      \centering
      \framebox{\parbox{3.2in}{
      \includegraphics[scale=.5]{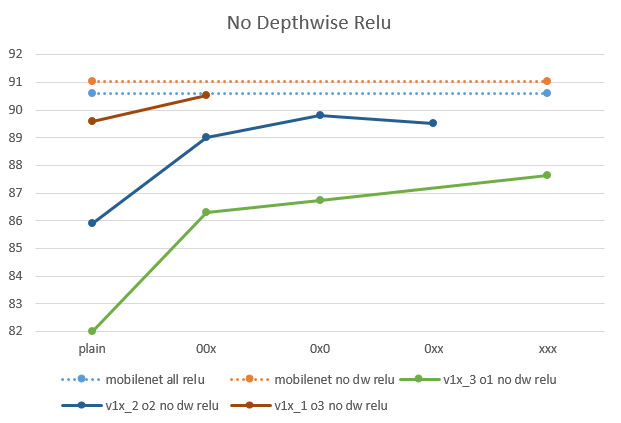}
      }}
      \label{nodwrelufig}
      \caption{}
\end{figure}

Table \ref{xxnet_no_dw_relu} and Figure \ref{nodwrelufig} show that removing a relu from the depthwise layer increases the accuracy in all of the networks, with the lowest being 82.01. It shows that we can match accuracy for the basline mobilenet with $>6.6x$ fewer parameters in the convolutional layers. The relu then has the effect of removing the influence of the inputs to the layer on being able to be expressed at each of the outputs.

\begin{table}[ht]
\caption{xxnet accuracy on cifar-10 with only end relu}
\label{xxnet_end_relu}
\begin{center}
\begin{tabular}{|c|c|c|c|c|c|c|}
\hline
& Shuffles & 000 & 00x & 0x0 & 0xx & xxx\\
\hline
\multirow{3}{*}{\shortstack{Major\\ Network}} & v1x\_3 o1 & 84.28 & 88.64 & 89.68 & N/A & 91.03 \\
\cline{2-7}
& v1x\_2 o2 & 87.71 & 89.62 & 90.67 & 90.21 & N/A \\
\cline{2-7}
& v1x\_1 o3 & N/A & N/A & N/A & N/A & N/A \\
\hline
\end{tabular}
\end{center}
\end{table}

\begin{figure}[thpb]
      \centering
      \framebox{\parbox{3.2in}{
      \includegraphics[scale=0.5]{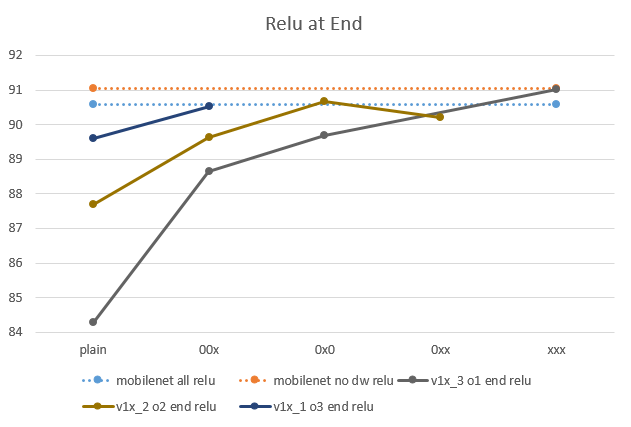}
      }}
      \label{endrelufig}
      \caption{}
\end{figure}

As shown in Table \ref{xxnet_end_relu} and Figure \ref{endrelufig} with the relu at the end of the network and none in between, we see that the network with two pointwise mini-layers is able to match and exceed the baseline mobilenet when shuffled in the correct place. The network with three pointwise mini-layers reaches the accuracy of mobilenet without a relu in the depthwise mini-layer with $>6.6x$ fewer parameters in the convolutional layers. 

Additionally Figure 4 shows that while mobilenet had stopped increasing in accuracy, xxnet v1x\_3 o1 was still gaining accuracy at a much higher rate.

\section{Future Work}
As mentioned there are more variations in the group size and group count for these xxnet networks as well as better shuffling in growth layers. With more experimentation and time to train the ImageNet dataset would be a good comparison point with Mobilenet v1.

\section{Conclusion}
Our results suggest that xxnet can get similar accuracy for far fewer parameters than Mobilenet v1. It also shows that ShuffleNet was not able to produce equivalent accuracy without adding in a new feature like residual layers because the relu affected the expression of the inputs at the outputs of each layer. This research shows that by applying a networking concept like the clos network separable networks with more layers can help explore networks with fewer parameters without sacrificing accuracy. As these ideas are further explored, convolutional networks will have less memory and computational requirements which will help enable the future of automation in cars, drones, and cities.

\addtolength{\textheight}{-12cm}   



\section*{APPENDIX}

Appendices


\end{document}